\documentclass[sigconf]{acmart}

\copyrightyear{2020}
\acmYear{2020}
\setcopyright{acmcopyright}\acmConference[KDD '20]{Proceedings of the 26th ACM SIGKDD Conference on Knowledge Discovery and Data Mining}{August 23--27, 2020}{Virtual Event, CA, USA}
\acmBooktitle{Proceedings of the 26th ACM SIGKDD Conference on Knowledge Discovery and Data Mining (KDD '20), August 23--27, 2020, Virtual Event, CA, USA}
\acmPrice{15.00}
\acmDOI{10.1145/3394486.3403219}
\acmISBN{978-1-4503-7998-4/20/08}

\usepackage{enumerate}
\usepackage{multirow}
\usepackage{subfiles}
\usepackage{graphicx}
\usepackage{booktabs}
\usepackage{ifthen}
\usepackage{tikz}

\usetikzlibrary{calc}
\usetikzlibrary{shapes.misc, positioning, shapes.geometric, arrows}

\usepackage[subtle]{savetrees}

\DeclareGraphicsExtensions{{.pdf, .png, .jpg}}
\graphicspath{ {images/} }

\newcommand{\xhdr}[1]{\vspace{3mm}\noindent{{\bf #1.}}}

\begin{document}

\title{Aligning~Superhuman~AI~with~Human~Behavior: Chess~as~a~Model~System}
	
\author[R. McIlroy-Young]{Reid McIlroy-Young}
\email{reidmcy@cs.toronto.edu}
\affiliation{%
  \institution{Department of Computer Science\\University of Toronto}
}

\author[S. Sen]{Siddhartha Sen}
\email{sidsen@microsoft.com}
\affiliation{%
  \institution{Microsoft Research}
}

\author[J. Kleinberg]{Jon Kleinberg}
\email{kleinberg@cornell.edu}
\affiliation{%
  \institution{Department of Computer Science\\
  Cornell University}
}

\author[A. Anderson]{Ashton Anderson}
\email{ashton@cs.toronto.edu}
\affiliation{%
  \institution{Department of Computer Science\\University of Toronto}
}
	
	\begin{abstract}
		As artificial intelligence becomes increasingly intelligent---in some cases, achieving superhuman performance---there is growing potential for humans to learn from and collaborate with algorithms. 
However, the ways in which AI systems approach problems are often different from the ways people do, and thus may be uninterpretable and hard to learn from.  
A crucial step in bridging this gap between human and artificial intelligence is modeling the granular actions that constitute human behavior, rather than simply matching aggregate human performance.

We pursue this goal in a model system with a long history in artificial intelligence: chess. 
The aggregate performance of a chess player unfolds as they make decisions over the course of a game. 
The hundreds of millions of games played online by players at every skill level form a rich source of data in which these decisions, and their exact context, are recorded in minute detail. Applying existing chess engines to this data, including an open-source implementation of AlphaZero, we find that they do not predict human moves well. 

We develop and introduce Maia, a customized version of Alpha\-Zero trained on human chess games, that predicts human moves at a much higher accuracy than existing engines, and can achieve maximum accuracy when predicting decisions made by players at a specific skill level in a tuneable way. 
For a dual task of predicting whether a human will make a large mistake on the next move, we develop a deep neural network that significantly outperforms competitive baselines. 
Taken together, our results suggest that there is substantial promise in designing artificial intelligence systems with human collaboration in mind by first accurately modeling granular human decision-making.

	\end{abstract}	

	
	
	
	%thumbnail
	%Our deep learning model, Maia, reading a chess board. Maia can predict the exact move a human will make in over 50% of chess positions.

	\addtolength{\abovecaptionskip}{-5pt}
	\addtolength{\textfloatsep}{-15pt}
	
	\maketitle
    
	\section{Introduction}

\newcommand{\omt}[1]{}

Artificial intelligence is becoming increasingly intelligent, equalling and surpassing peak human performance in an increasing range of domains~\cite{esteva2017dermatologist,silver2018general}.  In some areas,
once algorithms surpass human performance, people will likely stop
performing tasks themselves (e.g. solving large systems of equations).  
But there are many reasons why other domains
will continue to see a combination of human and AI participation
even after AI exceeds human performance---either because of
long transitional periods during which people and
algorithms collaborate; or due to 
regulations requiring human oversight for important decisions; or because
people inherently enjoy them.  
In such domains, there are rich opportunities for 
well-designed algorithms to assist, inform, or teach humans.
The central challenge in realizing these opportunities is that algorithms
approach problems very differently from the ways people do,
and thus may be uninterpretable, hard to learn from, or even
dangerous for humans to follow.

A basic step in these longer-range goals is thus to develop AI techniques
that help reduce the gap between human and algorithmic approaches
to solving problems in a given domain.
This is a genre of problem that is distinct from maximizing the performance
of AI against an absolute standard; instead, it asks whether we can 
create algorithms that more closely approximate human performance---using
fidelity to human output as the target rather than an
absolute standard of ground truth.
This type of question has begun to arise in a number of domains
where human specialists with deep expertise engage in 
decision-making with high stakes, for applications such as
medicine, law, hiring, or lending \cite{kleinberg2018human}.
But it remains difficult even to define the question precisely in general.
Approximating human performance should not simply mean matching 
 overall performance; a human and an AI system
performing the same classification task with comparable levels of accuracy
might nonetheless be making extremely different decisions
on those cases where they disagree.
The crucial question is whether we can create AI systems that
approximately emulate human performance on an instance-by-instance basis,
implicitly modeling human {\em behavior} rather than simply
matching aggregate human performance.
Moreover, the granularity matters---each instance of a complex task
tends to involve a sequence of individual judgments, and aligning
human and AI behavior benefits from performing the alignment at
the most fine-grained level available.

Furthermore, as human ability in domains varies widely, we want systems that can emulate human behavior at different levels of expertise in a tuneable way.   
However, we currently fail to understand crucial dimensions of this question.
In particular, consider a domain in which the strongest AI systems
significantly outperform the best human experts.
Such AI systems tend to have natural one-dimensional parameterizations
along which their performance monotonically increases---for example, we can
vary the amount of training data in the case of data-driven classification,
or the amount of computation time or search depth in the case of combinatorial search.
It is natural, therefore, to consider attenuating the system along this
one-dimensional path---e.g., systematically reducing the amount of
training data or computation time---to successively match different
levels of aggregate human performance.
We sometimes imagine that this might make the system more human-like
in its behavior; but in fact, there is no particular reason why this
needs to be the case.
In fact, we have very little understanding of this fundamental
{\em attenuation problem}: do simple syntactic ways of reducing an
AI system's performance bring it into closer alignment with
human behavior, or do they move it further away?
And if these simple methods do not achieve the desired alignment,
can we find a more complex but more principled parametrization of
the AI system, such that successive waypoints along this
parametrization match the detailed behavior of different levels
of human expertise?

\xhdr{A Model System with Fine-Grained Human Behavior}
In this work, we undertake a systematic study of these issues
in a model system with the necessary algorithmic capabilities and
data on human decision-making to fully explore the underlying questions.
What are the basic ingredients we need from such a model system?
\begin{enumerate}[(i)]
\item It should consist of a task for which AI has achieved superhuman performance, so that an AI system at least has the potential to match the full range of human skill without running into constraints imposed by its own performance.

\item There should be a large number of instances in which the context of each human decision and which action the human chose is recorded in as much fine-grained detail as possible.
\item These instances should include decisions made by people from a wide range of expertise, so that we can evaluate the alignment of an
AI system to many different levels of human skill.
\end{enumerate}

In this work, we use human chess-playing as our model system.
Let us verify how the domain meets our desiderata. 
Programming a computer to play chess at a high level was a long-standing holy grail of artificial intelligence, and superhuman play was definitively achieved by 2005 (point (i)). 
Humans have nonetheless continued to play chess in ever-increasing numbers, playing over one billion games online in 2019 alone. 
The positions players faced, the moves they made, and the amounts of time they took to play each move are digitally recorded and available as input to machine learning systems (point (ii)). 
Finally, chess is instrumented by a highly accurate rating system that measures the skill of each player, and chess admits one of the widest ranges of skill between total amateurs and world champions of any game (point (iii)). %CITE

What does it mean to accurately model granular human behavior in chess? 
We take a dual approach. 
First and foremost, we aim to be able to predict the decisions people make during the course of a game. 
This stands in contrast with mainstream research in computer chess, where the goal is to algorithmically play moves that are most likely to lead to victory.  
Thus, given a position, instead of asking ``What move should be played?'', we are asking, ``What move will a human play?''.
Furthermore, following our motivation of producing AI systems that
can align their behavior to humans at many different levels of skill,
we aim to be able to accurately predict moves made by players from a wide variety of skill levels.
This refines our question to: ``What move will a human at this skill level play?''.

Secondly, we aim to be able to predict when chess players will make 
a significant mistake. 
An algorithmic agent with an understanding of when humans of various levels are likely to go wrong would clearly be a valuable guide 
to human partners.

% stockfish doesn't work
Chess is also a domain in which the process of
{\em attenuating} powerful algorithms has been extensively studied.
As chess engines became stronger, playing against them became less fun and instructive for people. 
In response to this, online chess platforms and enthusiasts started to develop weaker engines so that people could play them and stand a fighting chance. 
The most natural method, which continues to be the main technique today, is limiting the depth of the game tree that engines are allowed to search, effectively imposing ``bounded rationality'' constraints on chess engines.
But does this standard method of attenuation produce better alignment
with human behavior?
That is, does a chess engine that has been weakened in this way do
a better job of predicting human moves?

Anecdotally, there is a sense among chess players that although
weaker chess engines are (clearly) easier to defeat, they do not 
necessarily seem more human-like in their play.
But there appears to be no quantitative empirical evidence one way or
the other for even this most basic question.
Thus, to establish a baseline, we begin in the subsequent sections by 
testing whether depth-limited versions of Stockfish, the reigning computer world chess champion, successfully predicts what humans of various strengths will play. 
Specifically,
we train 15 versions of Stockfish, each limited to searching the game tree up to a specific depth, and we test its accuracy in predicting which
moves will be made by humans of various ratings. 
We find that each version of Stockfish has a prediction accuracy that rises monotonically with the skill level of the players it is predicting, implying that it is not specifically targeting and matching moves made by players of lower skill levels. 
Moreover, while there are some interesting non-monotonicities that we expose, we find that stronger versions of Stockfish are better than other versions
at predicting the moves of human players of almost all skill levels.
This is a key point: if your goal is to use Stockfish to predict the
moves of even a relatively weak human player, you will achieve the best performance
by choosing the strongest version of Stockfish you can, despite the
fact that this version of Stockfish plays incomparably better chess
than the human it is trying to predict.
Depth-limiting classic chess engines thus does not pass our test of
accurately modeling granular human behavior in chess, in line with the
anecdotal experience of chess players.

% leela doesn't work
In 2016, DeepMind revolutionized algorithmic game-playing with the introduction of a sequence of deep reinforcement learning frameworks culminating in AlphaZero, an algorithm that achieved superhuman performance in chess, shogi, and go by learning from self-play. 
This led to much excitement in the chess world, not only for its unprecedented strength, but also for its qualitatively different style of play. 
In particular many commentators pointed out that AlphaZero, and its open-source implementation counterpart Leela, played in a more ``human'' style. 
Furthermore, these methods have a natural one-dimensional pathway
for attenuation, simply by stopping the self-play training early. 
To test whether this form of attenuation does better at aligning
with human behavior, we performed the same type of test with Leela
that we did with Stockfish.
We found that although Leela matches human moves with greater accuracy than depth-limited Stockfish, its accuracy is relatively constant across the range of player skill levels, implying that any one version of Leela isn't specifically targeting and modeling a given skill level. 
Thus, using AlphaZero-style deep reinforcement learning off the shelf does not pass our test of accurately modeling granular human behavior either.

\xhdr{Aligning Algorithmic and Human Behavior}
% maia
Having established the limitations of existing approaches,
we now discuss our results in aligning algorithmic
and human behavior.
As we show, we achieve performance in our model system that is
qualitatively different than the state-of-the-art; and in so doing,
we hope to suggest a road map for how similar efforts in other
domains might proceed.

In our approach, we repurpose the AlphaZero deep neural network framework to predict human actions, rather than the most likely winning move.
First, instead of training on self-play games, we train on human games recorded in datasets of online human play.
This encourages the neural net's
policy network to learn moves that humans are likely to play. 
Second, and crucially, 
to make a move prediction, we do not conduct any tree search---the policy network is solely responsible for the prediction. 
Third, we make further architectural changes to maximize predictive performance, including incorporating the past history of the game.
The resulting model, which we call {\em Maia}, is then tested
on its ability to predict human moves.
We find first of all that Maia achieves much higher move prediction 
accuracy than either Stockfish or Leela.
But Maia also displays a type of behavior that is qualitatively 
different from these traditional chess engines:
it has a natural parametrization under which it can be targeted
to predict human moves at a particular skill level.
Specifically, we show how to train nine separate Maia models, one for each bin of games played by humans 
of a fixed (discretized) skill level, and we find that the move
prediction performance is strikingly unimodal, with each Maia model
peaking in performance near the rating bin it was trained on. 
This is fundamentally different from the parametrized behavior of
either standard chess engines like Stockfish or neural network engines
like Leela: by developing methods attuned to the 
task of modeling granular human decisions, we can achieve high levels of accuracy at this problem, and can target to a specific human skill level.

% blunder prediction
In our second main focus for the paper, 
we turn to predicting whether human players will make 
a significant mistake on the next move, often called a \emph{blunder}. 
For this, we design a custom deep residual neural network architecture and train on the same data. 
We find that this network significantly outperforms competitive baselines at predicting whether humans will err. 
We then design a second task in which we restrict our attention to the most popular positions in our data---those that hundreds of people have faced---and aim to predict whether a significant fraction of the population faced with a given decision will make a mistake on it. 
Again, our deep residual network outperforms competitive baselines. 

Taken together, our results suggest that there is substantial promise in designing artificial intelligence systems with human collaboration in mind by first accurately modeling granular human decision-making.

	\section{Related Work}

\newcommand{\omt}[1]{}

Our work connects to several relevant literatures.
First, the contrasts between the approaches to problems
taken by humans and AI motivates the study of interpretability,
which seeks to define notions of understandability or explainability
for AI systems \cite{doshi-velez-science-interp,lipton-mythos-interp,zeng-ustun-interpretable}.
Our approach here is motivated by similar issues, but seeks to
design AI systems whose observable behavior (through the chess moves
they make) is more directly aligned with the observable behavior of
humans in the domain.
There are also connections to {\em imitation learning}
\cite{stadie-imitation}, in which AI systems learn by observing
human actions; the emphasis in this line of work tends to be on
settings where humans outperform AI, whereas we are interested in
using AI to align with human behavior in settings where humans
have significantly weaker performance.

In human-computer interaction, human factors research,
and related areas, there has been interest in systems
that could build models of human decision-making, so as to provide
more effective forms of assistance \cite{horvitz-mixed-initiative},
reductions in human error rate 
\cite{kirwan-human-reliability,salvendy-human-error},
and targeted educational content \cite{anderson1985intelligent}.

Finally, chess has long been used as a model system for both
artificial intelligence \cite{mccarthy-chess-drosophila}
and cognitive psychology \cite{charness-psych-chess-review}, and 
recent work has used powerful chess engines and chess information resources
to probe human decisions and errors
\cite{anderson2017assessing,biswas-regan-icmla15}.
Our work uses chess to focus on a crucial but poorly understood issue,
the possibilities to align algorithmic approaches with human ones.

	\section{Data and Background}

Our work leverages two radically different chess engines: Stockfish and Leela. Stockfish is a traditional chess engine that uses heuristic functions to evaluate positions, combined with a classical alpha-beta game tree search. Its evaluation function uses a set of rules developed by human experts that assign values to the pieces and their positional relationships in \emph{centipawn} units, or cp (the positional equivalent of 1 pawn equals 100 centipawns). We use Stockfish's evaluations to estimate the win probability of humans from certain board positions in our move matching analysis (Section \ref{delta_sec}), as well as to quantify the mistakes made by humans in our blunder prediction analysis (Section \ref{blunder_sec}). In contrast, Leela, an open-source implementation of AlphaZero, uses a deep neural network to approximate the value of each position, combined with a Monte Carlo tree search (a type of reinforcement learning). We repurpose Leela's neural network architecture for our move matching task, but instead of learning from self-play games, we learn from real human games. 

Throughout our work, we use chess games played by humans on \href{https://lichess.org}{Lichess.org}.
Lichess is a popular, free, open-source chess platform, on which people have played over 1 billion games at a current rate of over 1 million games per day. 
These games are played live at quick time controls between human players of all skill levels, ranging from total amateurs to the current world champion, Magnus Carlsen. 
Lichess has a very active developer and support community, and is known to have some of the most advanced measures in place to remove cheaters and other bad actors. As a result, the games database is virtually entirely composed of genuine human games. 

During a regular game on Lichess, both players start with an agreed-upon amount of time on their clock, typically between 3 and 10 minutes each for the entire game, and this time ticks down when it is their turn to move. 
If a player runs out of time, they lose the game. 
Games are clustered into different formats corresponding to how fast they are, including HyperBullet (30 seconds per player), Bullet (1 minute per player), Blitz (3--8 minutes per player), Rapid (8--15 minutes per player), and Classical (longer). 
In this paper, we ignore games played at the fastest time controls, HyperBullet and Bullet, since players are often more focused on not losing on time than playing quality moves. 

Every player on Lichess has a separate chess rating for each format listed above. A rating is a single number that represents a player's skill, with higher values indicating higher skill. 
On Lichess, the rating system used is Glicko-2, which is a descendant of the ubiquitous Elo rating system. 
A player's rating goes up or down after every game, depending on if they won or lost, and is updated intuitively: when a player beats a much higher-rated opponent, their rating increases more than when they beat a much lower-rated opponent.
The mean rating on Lichess is 1525, with 5\% of people below 1000 and 10\% above 2000. 
	\section{Move-Matching}

We now turn to our main task, developing an algorithm that can accurately predict the moves chosen by human chess players of various skill levels. In this way, we can begin to build models of granular human behavior that are targeted to specific skill levels.

\subsection{Task setup}
\label{mmsetup}

One advantage of operating in a model system, as we are, is that one can formulate the fundamental tasks in a very clean manner. Given a chess position that occurred in a game between two human players, we want to correctly predict the chess move that the player to move played in the game. Move-matching is thus a classification task, and any model or algorithm that takes a chess position as input and outputs a move has some performance on this task.

Since one of our main goals in this work is to develop an algorithm that can mimic human behavior at a specific skill level, we need to design an evaluation scheme that can properly test this.
To this end, we create a collection of test sets for evaluation, one for each narrow rating range.
First, we create rating bins for each range of 100 rating points (e.g.\ 1200-1299, 1300-1399, and so on).
We collect all games where both players are in the same rating range, and assign each such game to the appropriate bin.
We create 9 test sets, one for each bin for the 9 rating ranges between 1100 and 1900 (inclusive).
For each test set, we draw 10,000 games from its corresponding bin, ignoring Bullet and HyperBullet games. Within each game in a test set, we discard the first 10 ply (a single move made by one player is one ``ply'') to ignore most memorized opening moves, and we discard any move where the player had less than 30 seconds to complete the rest of the game (to avoid situations where players are making random moves).
After these restrictions, each test set contains roughly 500,000 positions each.

We will evaluate all models and engines with these 9 test sets, generating a prediction curve for each one.
This curve will tell us how accurately a model is predicting human behavior as a function of skill level. Let us reflect on how successfully modeling granular human behavior would manifest in this curve.

First, we want high move-matching accuracy --- correctly predicting which moves humans play is our primary goal.
But there's an important barrier to any model's performance on this task: a single model outputting a predicted move for each position by definition can't achieve near-perfect move-matching across all skill levels, since players at different skill levels play different moves (hence their difference in ability).
How then do we want a model's errors to be distributed?  This leads to a second desideratum after high move-matching accuracy:
we would like to have a parametrized family of models, ordered by increasing levels of human skill, such that the model in the family associated with skill $x$ would achieve maximum accuracy on players of skill $x$, with accuracy falling away on both sides of $x$.

% round down to nearest 100. both players in same bin
% 10,000 games from each bin.
% no bullet and hyperbullet
% in each game, take out first 10 ply and take out time pressure: if you have less than 30 seconds, skip (neither of these have an effect)
% all from december 2019
%

%A human like engine can be considered to be classifier, it is taking in the board and outputting the next move a human would make. By framing it like this we can treat our problem as one of classification. To operationalize this further we consider the mean accuracy per game, instead of the accuracy per move. This is equivalent to a weighted average over the moves with long endgame moves in particular being down weighted. The testing set is comprised of 100,000 games in each rating bin. Each bin is 100 wide,  i.e. 1500 to 1599 is the 1500 bin, and for all games in each bin both players are within the range. As we consider the player's move separately that gives 200,000 data points per bin.

\subsection{Evaluating chess engines}

As mentioned in the Introduction, developing chess-playing models that capture human behavior is not a new problem.
As chess engines became stronger and stronger, eventually overtaking the best human players in the world, playing against them became less fun and less instructive.
In order to create more useful algorithmic sparring partners, online chess platforms and enthusiasts began altering existing chess engines to weaken their playing strength.
The most common method was, and continues to be, limiting the depth of the game tree that the engines are allowed to search.
This attenuation is successful in weakening engines enough for humans of most skill levels to be able to compete with them.
On Lichess, for example, the popular ``Play with the computer'' feature lets one play against 8 different versions of a chess engine, each of which is limited to a specific depth.
Weakening a chess engine does necessarily mean increasing its similarity to human behavior, but we should start by posing this as a question:
Since this method gives rise to a parameterized family of chess engines that smoothly vary in skill level (thus matching aggregate human performance at various levels), we ask whether these depth-limited engines already succeed at the move-matching task we've defined.  We will see next that they do not.

\xhdr{Stockfish}
Stockfish is a free and open-source chess engine that has won 5 of the last 6 computer chess world championships, and is the reigning world champion at the time of writing~\cite{wikitcec}.
Due to its strength and openness, it is one of the most popular engines in the world --- all computer analysis on Lichess, and the ``Play with the computer'' feature, is done using Stockfish.
We tested 15 depth-limited versions of Stockfish on the move-matching task, from depths 1--15, and report results on a subset of them for clarity.

\begin{figure}[t]
	\centering
	\includegraphics[width=.5\textwidth]{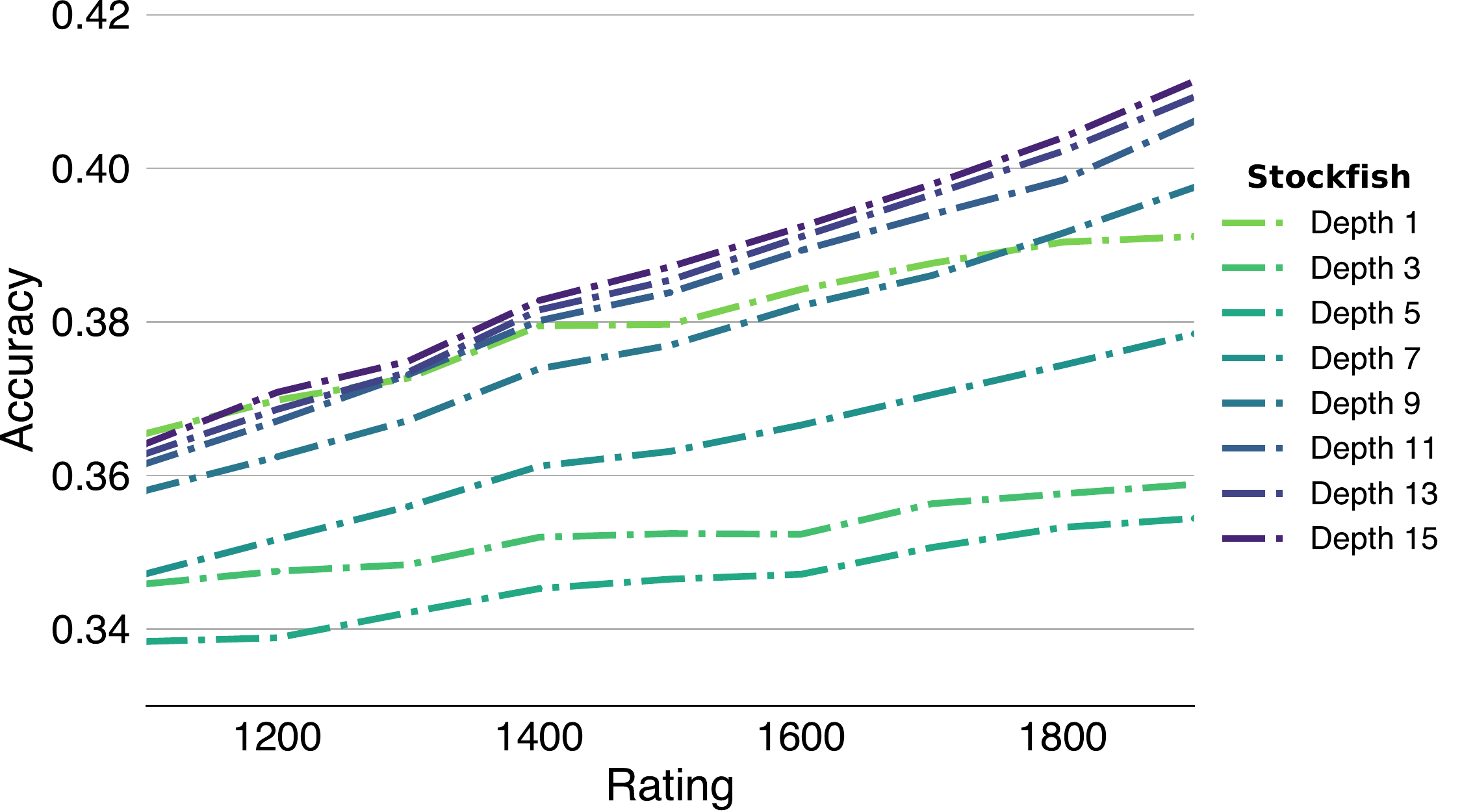}
	\caption{Move-matching performance of a family of depth limited Stockfish engines.}
	\label{move_matching_sf}
\end{figure}

Figure~\ref{move_matching_sf} shows the prediction accuracy curves for each engine in this subset.

First, we find that the Stockfish engines match human moves between 33--41\% of the time, establishing a baseline to compare other models against.
Furthermore, the accuracy is non-linear in the engine's depth limit: the Stockfish version limited at depth 1 matches human moves more often than the version limited at 5, but less often than the version limited at depth 15, which achieves the highest performance.
Most importantly, all of the curves are monotonically increasing in the strength of the human players being matched against; for example, depth 15 matches 1900-rated players 5 percentage points more than it matches 1100-rated players.
This implies that although depth-limited Stockfish engines are designed to play at lower levels, they are not playing moves that are more typical of weak players than strong players --- they do not target specific skill levels as we want.
Furthermore, the strongest versions we tested, such as depths 11, 13, and 15 shown in Figure~\ref{move_matching_sf}, have almost identical performance, despite the fact that they differ quite significantly in playing strength from each other.
This implies that as Stockfish increases (or decreases) in strength, it does so largely orthogonally from how humans increase or decrease in strength.
This is a clear demonstration that algorithmically matching aggregate human performance (winning and losing chess games) does not necessarily imply matching granular human actions (playing human moves).
To do so, we need another algorithmic approach.

\xhdr{Leela}
In 2016, Silver et al.\ revolutionized algorithmic game-playing with the introduction of a sequence of deep reinforcement learning frameworks culminating in AlphaZero, an algorithm that achieved superhuman performance in chess, shogi, and Go by learning from self-play~\cite{silver2017masteringgo,silver2018general}.
AlphaZero adopts a completely different approach to playing chess than classical chess engines such as Stockfish.
 Most engines derive their strength from conducting fast, highly optimized game tree searches incorporating techniques such as alpha-beta pruning, in combination with handcrafted evaluation functions that score any given chess position according to human-learned heuristics.
In comparison, AlphaZero learns from self-play games and trains a deep neural network to evaluate chess positions, and combines this with Monte Carlo Tree Search to explore the game tree.
This method is many orders of magnitude slower than classical game tree search methods, but the non-linear evaluation function is more flexible and powerful than the linear functions used by traditional engines.
AlphaZero's introduction came with the announcement that it crushed Stockfish in a 100-game match by a score of 28 wins to 0 with the rest drawn (although some later contested the experimental conditions of this match).
This led to much excitement in the chess world, not only for its unprecedented strength, but also for its qualitatively different style of play.
In particular, many commentators pointed out that AlphaZero played in a more dynamic, human style~\cite{kasparov2018chess}.
Furthermore, as the neural network slowly evolves from random moving to world-class strength, sampling it at various times in the training process is a natural way of producing a parameterized family of engines.

Given this context, we next test whether Leela Chess Zero, a free, distributed, open-source implementation of AlphaZero, succeeds at our move-matching task.
For this analysis, we select 8 different versions of Leela's 5th generation that significantly vary in strength.
Leela uses an internal rating system to rank its different versions, but we note that these ratings are not comparable to Lichess ratings. We refer to them as ``Leela ordinal ratings'' to emphasize this.
\begin{figure}[t]
	\centering
	\includegraphics[width=.5\textwidth]{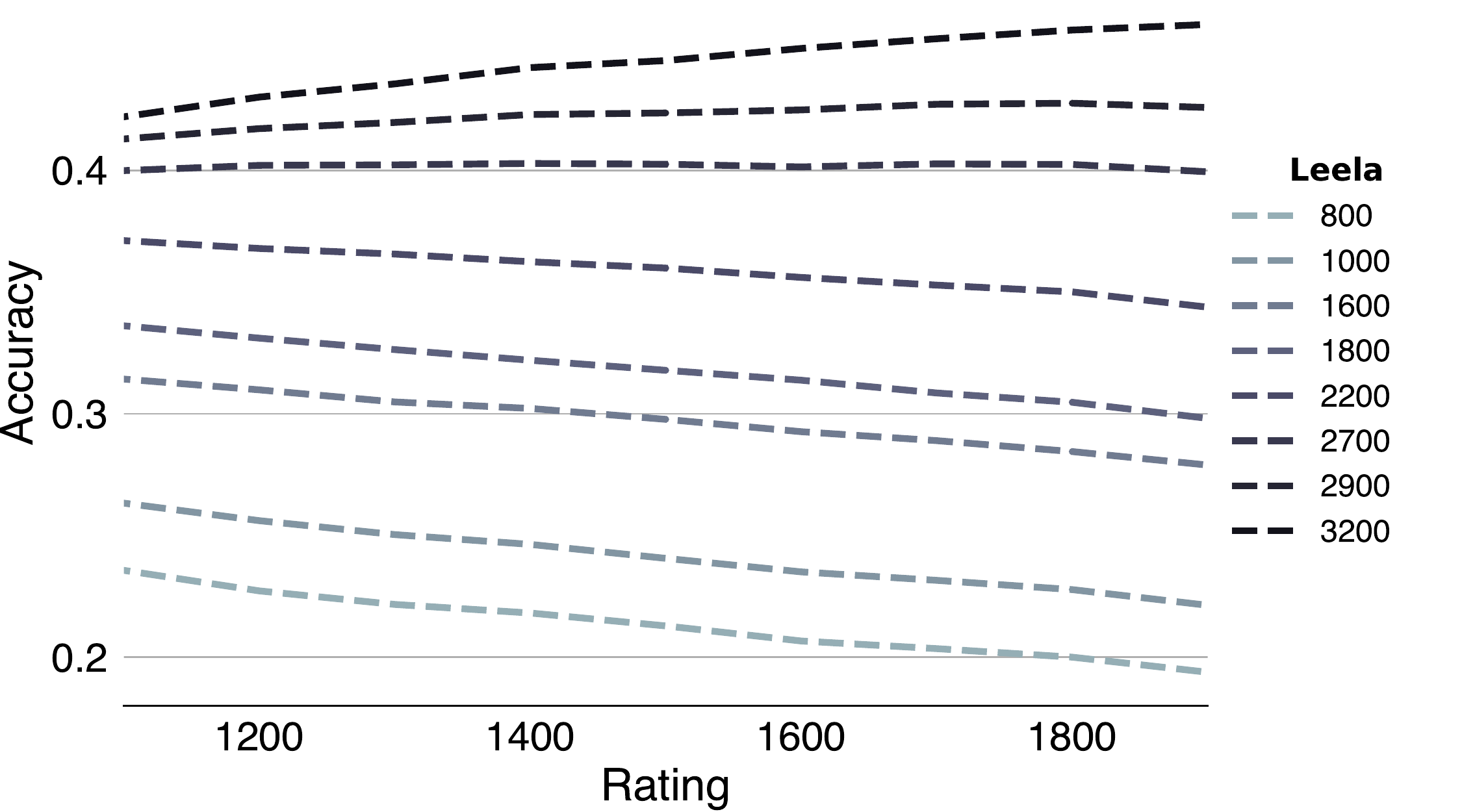}
	\caption{Move-matching performance of a family of Leela engines.}
	\label{move_matching_leela}
\end{figure}

Figure~\ref{move_matching_leela} shows the prediction curves of the family of Leela engines. The move-matching performance varies much more dramatically for this set than it does for Stockfish.
This is intuitive, as early versions of Leela are not that far removed from its initial random state, whereas later versions are incredibly strong.
We find that strong versions achieve higher move-matching accuracy than any version of Stockfish, scoring a maximum of 46\%.
However, all versions of Leela we tested have prediction curves that are essentially constant, or have a slight positive or slight negative slope.
Thus, even Leela does not match moves played by humans at a particular skill level significantly better than it matches moves played by any other skill level.
For example, Leela ordinal rating 2700 matches human moves 40\% of the time, no matter whether they are played by humans rated 1100, 1500, or 1900, and therefore it doesn't characterize human play at any particular level.
Neither traditional chess engines nor neural-network-based engines match human moves in a targeted way.
\vspace{-5pt}
\subsection{Maia}

\begin{figure}[t]
	\centering
	\includegraphics[width=.45\textwidth]{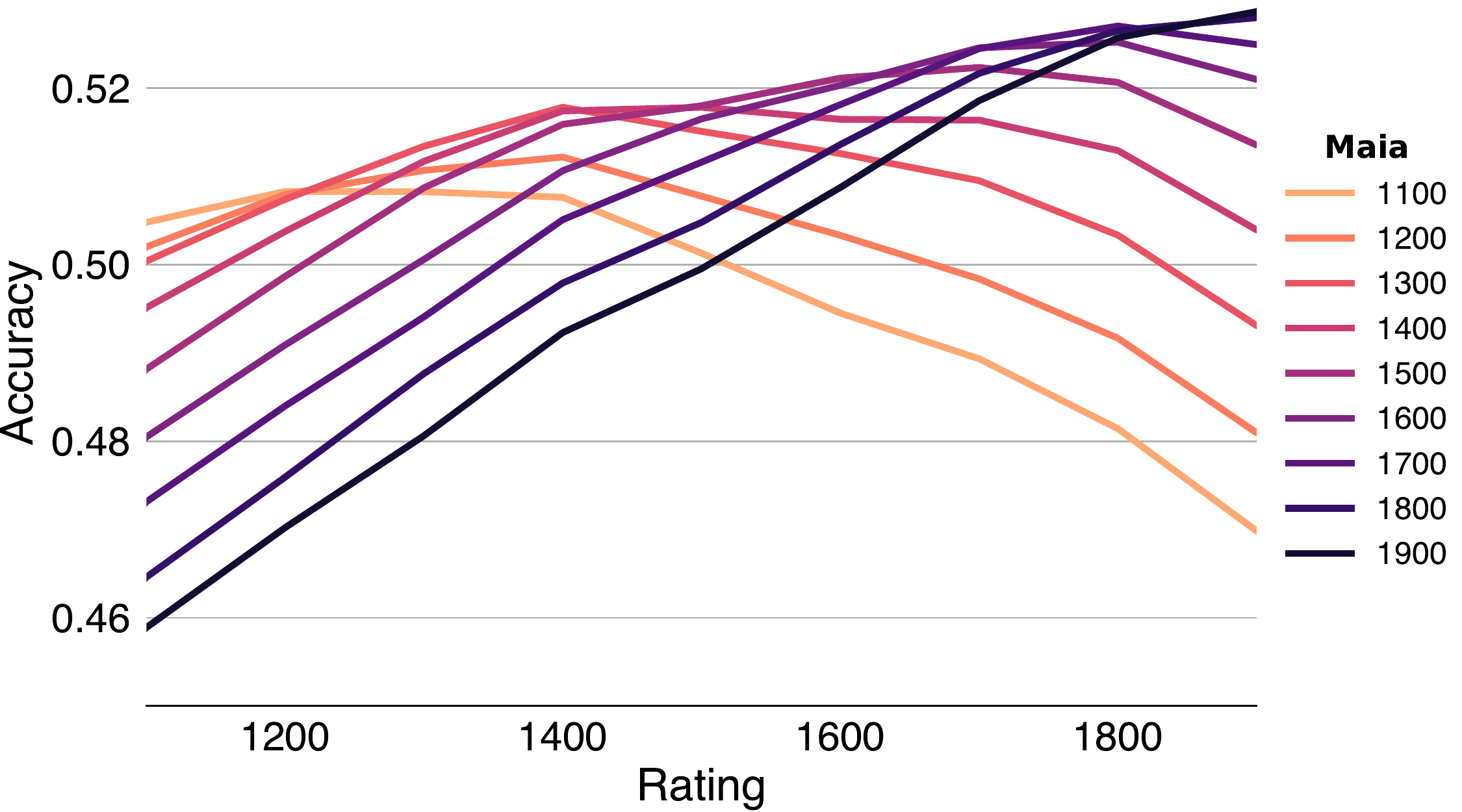}
	\caption{Move-matching performance of a family of Maia models.}
	\label{maia_lineplot}
\end{figure}

To accurately model granular human behavior in a tuneable way, we created a new deep neural network chess engine based on AlphaZero.
Since we are trying to explicitly model human behavior, our most fundamental change is to replace training on self-play games with training on human games.
By doing so, we are leveraging the billions of games played online as an incredibly rich resource from which we can learn and model human actions.
Whereas the policy head in AlphaZero's architecture tries to predict the best move, we tasked our neural network to predict the next move a human made.

Our second major distinction from AlphaZero is the following: while tree search is clearly important to strong chess-playing performance, we found that it degrades our performance at predicting human moves.  As a result, we do not conduct any tree search to make a move prediction.
At test time, we query the neural network with the current board position, which outputs the move it deems most likely.
In effect, we are asking what our representation sees as the most natural move to a human player, without explicitly calculating the ramifications of the move by searching the game tree.
We return to this, and further architectural decisions we made, later in this section.

\xhdr{Training} We constructed our training sets in a similar fashion to how we constructed our test sets (see Section~\ref{mmsetup}). This resulted in 9 training sets and 9 validation sets, one for each rating range between 1100 and 1900, with 12 million games and 120,000 games each, respectively (see the Appendix for full details).
We note that 12 million games is enough for Leela to go from random moving to a rating of 3000 (superhuman performance), and therefore far outstrips the usual training size for a single Leela network.

We then trained separate models, one for each rating bin, to predict the next move a human would make (policy) and the probability that the active player will win the game (value).
See the Supplement (section \ref{maia_sup}) for full implementation details, and the repository\footnote{\href{https://github.com/CSSLab/maia-chess}{https://github.com/CSSLab/maia-chess}} for the complete codebase.

\begin{figure}[t]
	\centering
	\includegraphics[width=.5\textwidth]{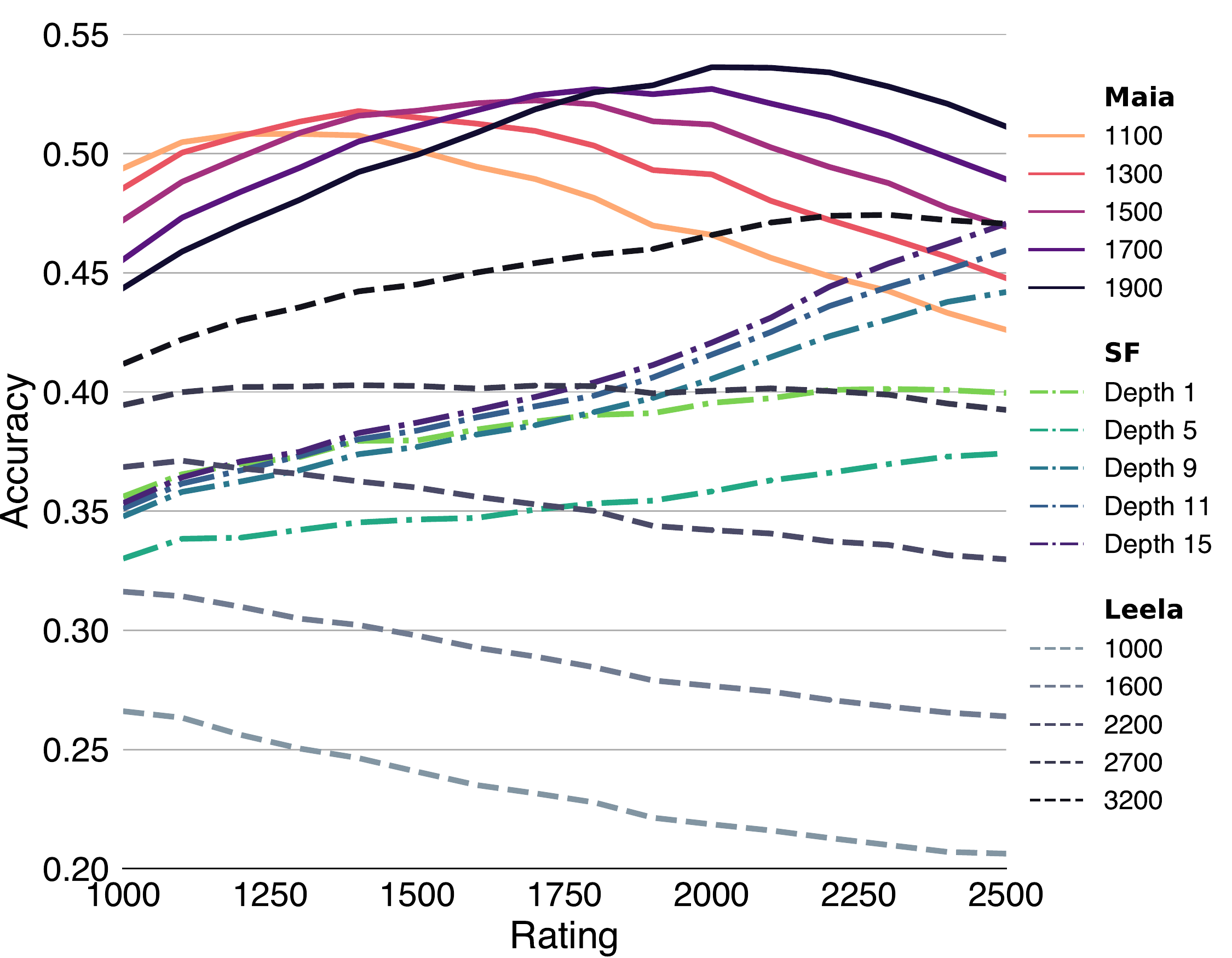}
	
	\caption{Comparison of move-matching performance for Maia, Stockfish, and Leela models.}
	\label{all_lineplot}
\end{figure}

\xhdr{Results} In Figure~\ref{all_lineplot}, we show how our models, which we call Maia, perform on the move-matching test sets.
Several important findings are apparent.
First, Maia models achieve high accuracy, far above the state-of-the-art chess engines discussed earlier.
The lowest accuracy, when Maia trained on 1900-rated players predicts moves made by 1100-rated players, is 46\% --- as high as the best performance achieved by any Stockfish or Leela model on any human skill level we tested.
Maia's highest accuracy is over 52\%.
Second, every model's prediction curve is strikingly unimodal.
Each model maximizes its predictive accuracy on a test rating range near the rating range it was trained on.
This means that each model best captures how players at a specific skill level play, and displays exactly the skill-level-targeting behavior that we are seeking.
Each model's predictive performance smoothly degrades as the test rating deviates further from the rating range it was trained on, which is both necessary to some degree, as humans of different levels make different moves, as well as desired, since it indicates the model is specific to a particular skill level.
Maia thus succeeds at capturing granular human behavior in a tuneable way that is qualitatively beyond both traditional engines and self-play neural network systems. %We emphasize that 

For ease of comparison, Figure~\ref{all_lineplot} overlays the move-matching performance for the Stockfish, Leela, and Maia families of engines.
We also include test sets that extend to rating ranges of 1000 and 2500 to show how predictive performance extends to the extreme ends of the skill distribution.
Over the rating ranges that Maia models were trained on (1100--1900), they  strictly dominate Leela and Stockfish: every Maia model outperforms every Leela and Stockfish model.
As the testing ratings increase after the last Maia training rating (1900), the Maia models degrade as expected, and the best Leela and Stockfish models begin to be competitive.
This is intuitive, since the strong engines are aligning better with strong humans, and the Maia models shown here were not correspondingly trained on humans of these strength levels. Maia 1900, however, still outperforms the best Stockfish and Leela models by 5 percentage points.

\xhdr{Further architectural decisions}
To achieve this qualitatively different behavior, we made two key architectural decisions.

\begin{figure}[t]
	\centering
	\includegraphics[width=.49\textwidth]{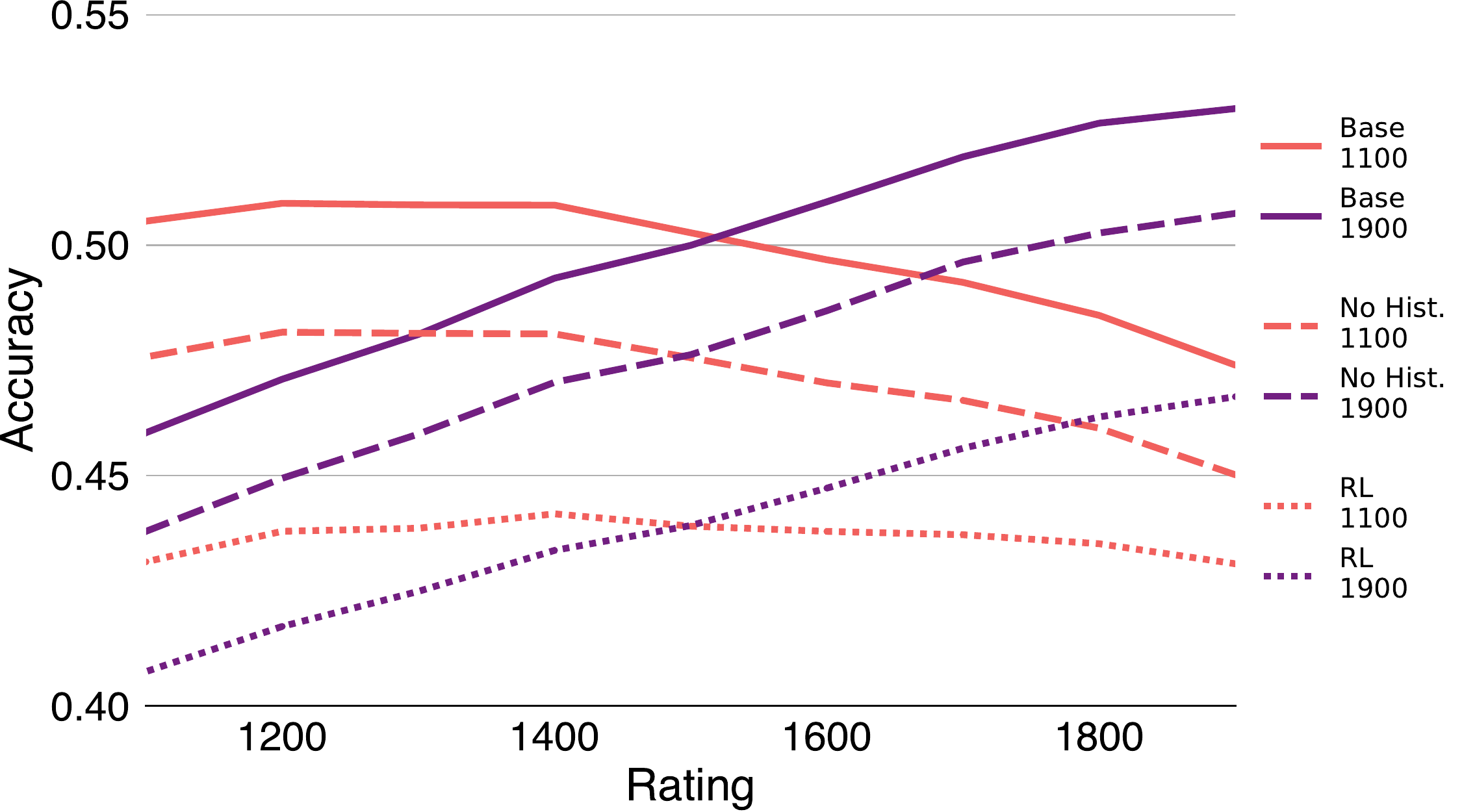}
	\caption{Move-matching performance of two Maia models, in either the base configuration, with no history provided, or with 10 rollouts of tree search performed.}
	\label{other_effects_lineplot}
\end{figure}

First, as mentioned above, we do not conduct any tree search to make a move prediction. Although Monte Carlo tree search is crucial to AlphaZero's strength and playing style, it tends to degrade move prediction performance in our setting. Here, we demonstrate this by comparing base Maia with a version of Maia that does 10 rollouts of tree search exploration (see \cite{silver2018general} for a detailed explanation).
Second, we give Maia the previous 12 ply (6 moves for each player) that were played leading up to the given position. We find that this significantly improves our move-matching accuracy.

The effects of these two architectural decisions are shown in Figure~\ref{other_effects_lineplot}. For all versions of Maia (here we show only those trained on 1100 and 1900 for clarity), including recent history and performing no tree search give large boosts in performance (3 percentage points and 5--10 percentage points, respectively). Increasing the number of rollouts does not affect performance.

\subsection{Model comparisons}\label{delta_sec}

We have established that Maia can model granular human decisions in a much more accurate and targeted way than existing engines.
Here, we compare the different families of models to better understand this relationship.

\xhdr{Model agreement}
How similar are each model's predictions?
Since we evaluated every model on the same test sets, we can compare their predictions and measure how often they agree.
For every pair of models, we measure how often they predict the same move in a given position across all positions in our test sets.

\begin{figure}[t]
	\centering
	\includegraphics[width=.46\textwidth]{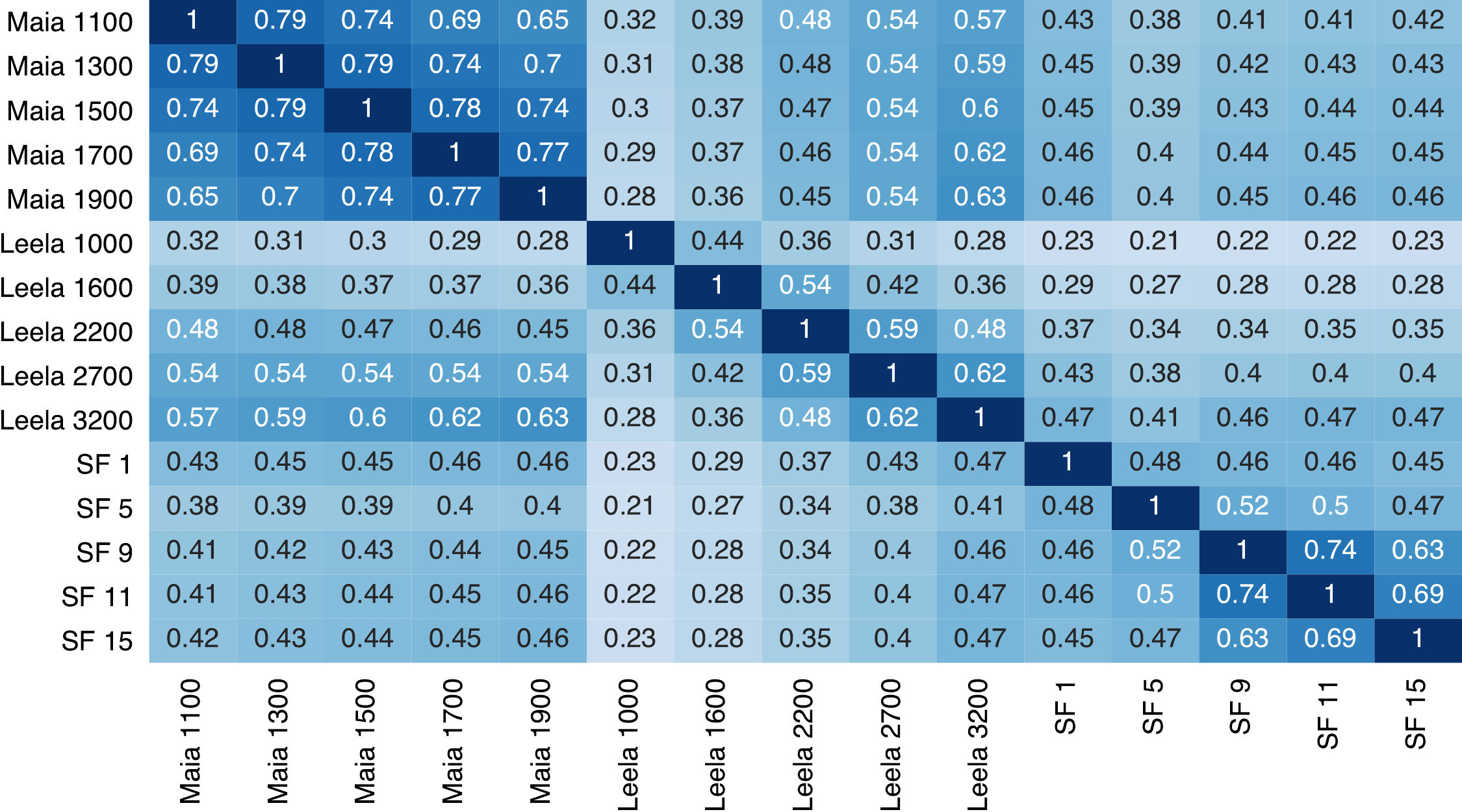}
	\caption{Per-move agreement between a selection of Maia, Leela, and Stockfish models.}
	\label{models_agreement}
	\vspace{10pt}
\end{figure}

These per-move agreement rates are shown in Figure~\ref{models_agreement}.
Most strikingly, the Maia models are very similar to each other, despite the very different skill levels they were trained on.
The highest inter-Maia agreement is 79\%, far higher than the agreement of any Leela or Stockfish model  with any other model.  The lowest inter-Maia agreement is 65\%, which is still higher than the agreement of any Maia model with a non-Maia model.
The Maia models therefore seem to predict moves that occupy a distinct ``subspace'', despite the differences in skill they are trained on.
In contrast, Leela and Stockfish agreement is lower and more dependent on similarity in strength. For example, Leela 2700 and 3200, and Stockfish depth 9 and depth 11, agree with each other at relatively high rates, but versions that are further apart in skill agree at very low rates.

\xhdr{Model maxima}
It is instructive to examine where each of the models achieve their best performance. In Table 1, we show each model, the test set on which it achieved its maximum performance, and its  accuracy on this test set. Each model family displays qualitatively different behavior. The Stockfish models all rise monotonically with test set rating, and thus achieve their best performance on the highest rating test set, 1900. Leela versions, on the other hand, first tend to monotonically decrease with rating, until they reach a certain strength, when the slope becomes monotonically increasing. The weaker models achieve their maximum on the lowest rated test set and the stronger models achieve it on the highest rated test set. (Leela 2700 is perfectly in the middle, achieving roughly the same performance across the full range.) Maia models, in contrast, achieve their maximum performance throughout the rating range, usually one rating range higher than their training set. Their maximum accuracy is consistently above 50\%.

\begin{table}[t]
    \centering
    \setlength{\tabcolsep}{2pt}
    \begin{tabular}{rllrllrll}
    \toprule
    \textbf{Maia}       & \textbf{Best}    &     \textbf{Acc.}  \hspace{10pt} &  \textbf{Leela} &  \textbf{Best}    &   \textbf{Acc.}   \hspace{10pt}  & \textbf{SF}       &  \textbf{Best}  & \textbf{Acc.}\\
    \midrule
    1100 & 1200 & 50.8\% & 1000 & 1100 & 26.3\% & 1  & 1900 & 39.1\% \\
    1300 & 1400 & 51.8\% & 1600 & 1100 & 31.4\% & 5 & 1900 & 35.4\% \\
    1500 & 1700 & 52.2\% & 2200 & 1100 & 37.1\% & 9 & 1900 & 39.8\% \\
    1700 & 1800 & 52.7\% & 2700 & 1400 & 40.2\% & 11 & 1900 & 40.6\% \\
    1900 & 1900 & 52.9\% & 3200 & 1900 & 46.0\% & 15 & 1900 & 41.1\% \\
    \bottomrule
    \end{tabular}
    \caption{Where selected Maia, Leela, and Stockfish models achieve their highest accuracy.}
    \label{top_elos}
\end{table}{}

\xhdr{Decision type}
In chess, as in other domains, decisions tend to vary in their complexity. Some chess positions only have one good move; in others there are many good moves. One way to measure this complexity is to consider the difference in quality between the best and second best moves in the position. If this difference is small there are at least two good moves to choose from, and if it's large there is only one good move. Here, we decompose our main results by calculating how Maia and Leela models match human moves, over all test sets, as a function of this difference.
We measure ``move quality'' by using Stockfish depth 15's evaluation function (the strongest engine of all we tested), then converting its evaluation into a probability of winning the game (see Supplement section \ref{cp_wr_supp} for more details).

\begin{figure}[t]
	\centering

	\includegraphics[width=.47\textwidth]{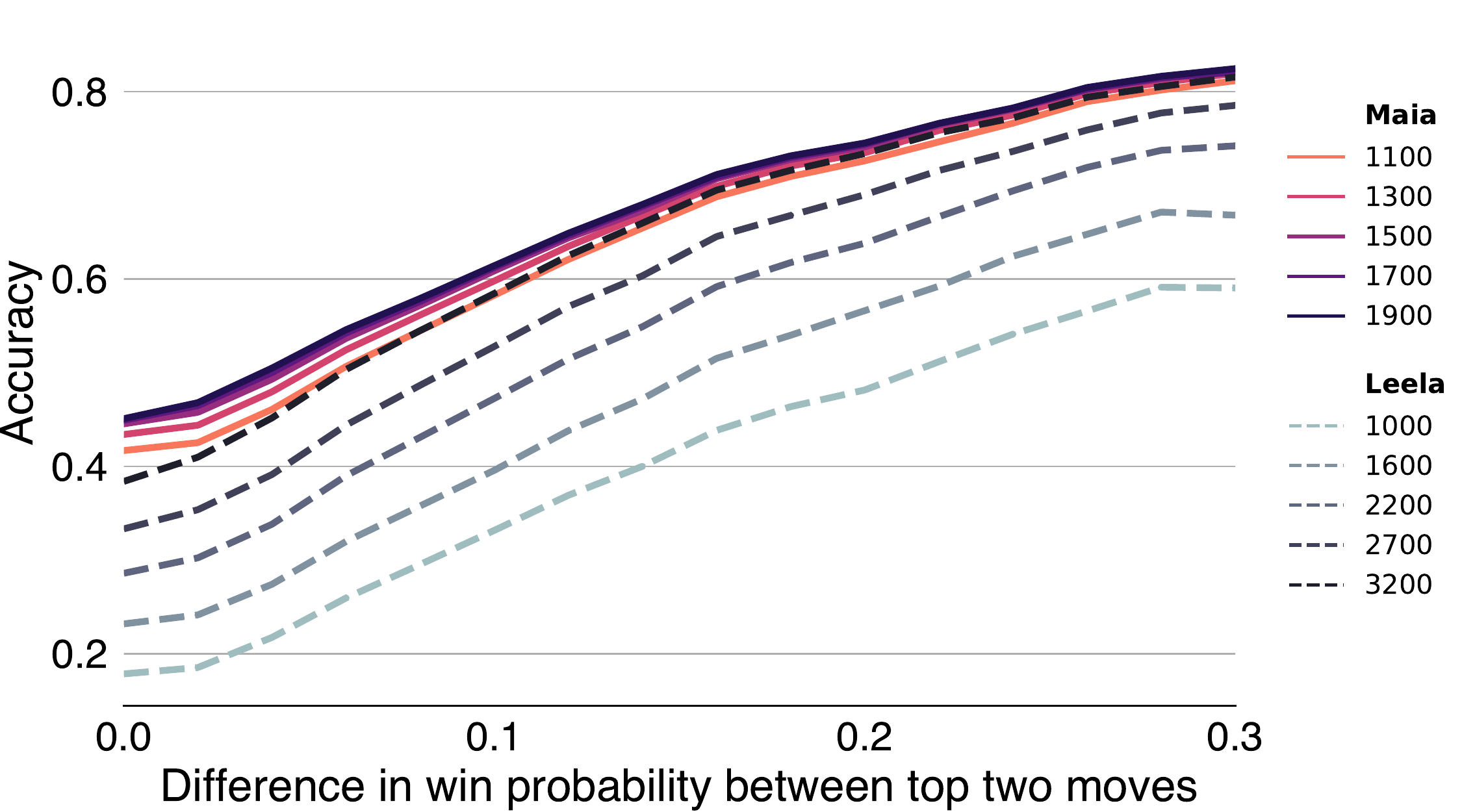}
	\caption{Move-matching performance as a function of the position complexity.}
	\label{delta_top2}
\end{figure}

In Figure~\ref{delta_top2}, we see that as the difference in quality between the two best moves increases, all models increase in accuracy. This makes sense --- in many positions with only one good move, the good move is an obvious one, such as recapturing a piece. More interestingly, Maia's improvement over Leela is relatively constant across the range; it is as much better at predicting non-obvious moves as it is predicting obvious moves in comparison with Leela.

\begin{figure}[t]
	\centering
	\includegraphics[width=.47\textwidth]{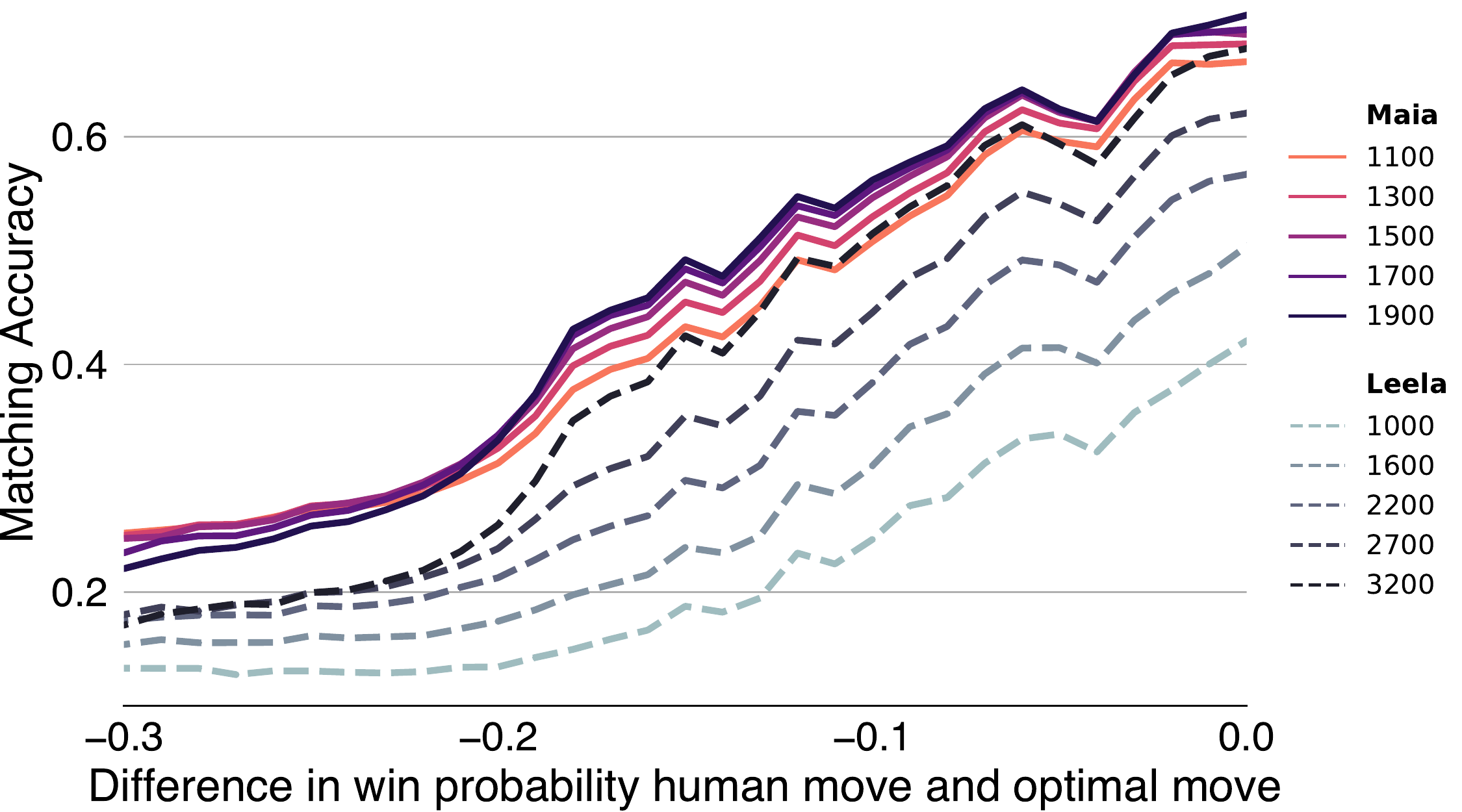}
	\caption{Move-matching performance as a function of the quality of the move played in the game.}
	\label{delta_human}
	\vspace{10pt}
\end{figure}

\xhdr{Decision quality}
Chess is also like other domains in that humans regularly make mistakes. Here, we analyze how move prediction performance varies with the quality of the human move. We again define move quality by translating Stockfish depth 15 evaluations into win probabilities, and measure the difference between the best move in the position and the move the human actually played. The size of this gap thus becomes a rough proxy for how good or bad the human move was. In Figure~\ref{delta_human}, we see that model performance tends to improve as the quality of the played move increases. This is intuitive, since good moves are easier to predict. Interestingly, every Maia model is almost always better than every Leela model across the entire range, including large errors (``blunders'', in chess terminology). Although blunders are undesirable and humans try to avoid them, Maia still does the best job of predicting them. To follow up on this, in the next Section we focus on predicting whether humans will blunder in a given position.

	\section{Predicting Errors}\label{blunder_sec}

We now turn to our second task of predicting whether a human facing a given chess position will make a significant mistake, or blunder. Strong performance on this genre of task would clearly be useful, as algorithmic systems could be equipped with an understanding of when humans might go wrong.

\subsection{Blunder Prediction}

\xhdr{Task}
Our task is to predict if the next move made by a human player will be a blunder. Instead of predicting the next human move, we can instead ask a more direct question: ``Is a human likely to blunder in the current position?'' Answering this question accurately can serve as a useful guide for human players, by suggesting whether a given position might be difficult for them.

As in Section~\ref{delta_sec}, we convert Stockfish's evaluation of a given position into a probability of winning the game (using a lookup table mapping Stockfish centipawn evaluations into the empirical probability of winning a game in our dataset; see the Supplement for more details). We then label a move as a blunder if it worsens a player's win probability by 10 percentage points or more.

We evaluated two formulations of the task. In the first ("board only"), models are only allowed to use the chess position as input.
In the second ("board and metadata"), models are provided metadata about the players and the game in addition to the board state. This metadata is limited to features that could be derived by a live observer of the game, such as the players' ratings, the percentage of time remaining for each player, and the cp score of the board. 

\xhdr{Data}
We used the same set of chess games as we did for the move prediction task, ending in September 2019 for the training set and using October 2019 as the test set.
Since some Lichess games are annotated with high-quality Stockfish evaluations (any player on Lichess can request this, e.g.\ for analysis or training), we restricted our attention to these games, which account for 10\% of all games.

To create the training set, we took all games with Stockfish evaluations and timing information from April 2017 to September 2019 and removed those played at fast time controls, as well as moves made with under 30 seconds remaining. We then classified each move as a blunder or non-blunder, and randomly down-sampled non-blunders until they were 150\% of the blunder set size.
The games in October 2019 were similarly processed and used as a balanced test set.
The final training set contains 182M blunders and 272M non-blunders; the test set contains 9M blunders and 9M non-blunders. During training, half of the batches were sampled from the blunder set and half were sampled from the non-blunder set.

\xhdr{Results}
We first evaluated a set of baselines, which we trained on a subset of moves from the full training set (100K moves from each month, or 3M total). We performed a grid search across hyperparameters and used a validation set to select the model with the highest AUC. Random forests models performed best in both task formulations, achieving 56.4\% accuracy when given just the board and 63\% accuracy when given the board state and metadata.

We then trained a fully connected neural network with 3 hidden layers that output a single number and used mean squared error as the loss function (more details of the network configuration and training appear in the Supplement). In both task formulations this model outperformed the best baseline.

Finally, we trained a residual CNN that we designed based on AlphaZero (full details are in the Supplement). When providing metadata to the model, we normalized the values to $[0,1]$ and supplied each value as an additional channel; this achieved better performance than feeding the metadata into a fully connected layer after the CNNs. This network's performance demonstrates a significant improvement over the fully connected network, achieving up to 71.7\% accuracy (see Table~\ref{blunder_results} for a summary of all the results).

In sum, we can also train AI systems to give us a better understanding of when people might go wrong.

\addtolength{\tabcolsep}{3pt}
\begin{table}[t]
    \centering
    \begin{tabular}{ccc}
        \toprule
         \textbf{Model} & \textbf{Board only} &  \textbf{Board and Metadata}\\
         \midrule
             Random Forest   & 56.4\% & 63.0\% \\
             Fully Connected &  62.9\% & 66.0\%\\
             Residual CNN         & 67.7\% & 71.7\% \\
         \bottomrule
    \end{tabular}
    \caption{Blunder prediction testing accuracy.}
    \label{blunder_results}
\end{table}

\subsection{Collective Blunder Prediction}

\xhdr{Task}
Throughout this paper we have mainly concerned ourselves with prediction tasks where the decision instances are mostly unique---chess positions usually occur once, and then never again.
But some positions recur many times.
We now focus on the most common positions in our data, and try to predict whether a significant fraction (>10\%) of the population who faced a given position blundered in it.

\xhdr{Data} We grouped individual decision instances by the position the player faced, normalizing them by standardizing color and discarding extraneous information (e.g.\ how many ply it took to arrive at the position), and counted how often each position occurred in our data. We kept all positions that occurred more than 10 times, which produced a training dataset with 1.06 million boards and a balanced test set with 119,000 boards. These boards are almost all from the early or latter stages of the game, since that is when exact positions tend to recur.

\xhdr{Results}
We trained fully connected, residual CNN, and deep residual CNN models. The 3-layer fully connected network achieves 69.5\% accuracy, whereas both the residual CNN and deep residual CNN perform even better (75.3\% and 76.9\%, respectively). This task is thus easier than individual blunder prediction, which may be due to the fact that grouping decisions together reduces noise. As we continue to bridge the gap between superhuman AI and human behavior in other domains, it will be interesting to contrast predicting individual decision quality with predicting how groups of people fare on similar or identical decisions, as we have done here.

	\section{Conclusion}

In an increasing number of domains, rapid progress on the goal of
superhuman AI performance has exposed a second, distinct goal:
producing AI systems that can be tuned to align with human behavior
at different levels of skill.
There is a lot we don't understand about achieving this goal,
including the relation between an AI system's absolute performance
and its success at matching fine-grained traces of human behavior.

We explore these questions in a setting that is particularly 
well-suited to analysis --- the behavior of human chess players
at a move-by-move level, and the development of chess algorithms
to match this human move-level behavior.
We begin from the finding that existing start-of-the-art chess algorithms
are ill-suited to this task: the natural ways of varying their strength
do not allow for targeting them to align with particular levels of
human skill.
In contrast we develop a set of new techniques, embodied in a new chess
model that we call {\em Maia}, which produces much higher rates of
alignment with human behavior; and more fundamentally, it is 
parametrized in such a way that it achieves maximum alignment at
a tuneable level of human skill.
In the paper, we have seen some of the design choices that lead
to this type of performance, and some of the implications for 
modeling different levels of human skill.
We also extend our methods to further tasks, including the 
prediction of human error.

There are a number of further directions suggested by this work.
First, it would be interesting to explore further where Maia's
improvements in human alignment are coming from, and whether we can
characterize subsets of the instances where additional techniques
are needed for stronger performance.
Potentially related to this is the question of whether an approach
like Maia can expose additional dimensions of human skill;
we currently rely on the one-dimensional rating scale, which
has proven to be quite robust in practice at categorizing chess players,
but with increasingly powerful approaches to fine-grained alignment,
we may begin identifying novel distinctions among human players
with the same rating.
And finally, it will be interesting to explore the use of these
techniques in an expanding collection of other domains, as we seek to model
and match fine-grained human behavior in high-stakes settings, on-line
settings, and interaction with the physical world.

\small{\xhdr{Acknowledgments} We thank the anonymous reviewers for helpful comments. AA was supported in part by an NSERC grant, a Microsoft Research Award, and a CFI grant. JK
was supported in part by a Simons Investigator Award, a
Vannevar Bush Faculty Fellowship, a MURI grant, and a
MacArthur Foundation grant.}
\vspace{-1pt}

	\bibliographystyle{ACM-Reference-Format}
	\bibliography{cites}
	\clearpage
	\section{Supplement}

Our full source code, data and models are available online\footnote{\href{https://github.com/CSSLab/maia-chess}{github.com/CSSLab/maia-chess}}. The raw data used are all downloaded from \href{https://database.lichess.org/}{database.lichess.org}, but converting the PGNs to files for our work is computationally expensive. Processing  April 2017 to December 2019 into a tabular form, one row per move, took about 4 days on a 160 thread 80 core server and used 2.5 TB of memory, and converting to the final format for our models another 3 days. So we also host some of those files too. We hope that the tabular data will be used in further work and have sought to include all the available data.

Additionally our move prediction models can be used as UCI chess engines and can be played against as the following bots on Lichess: \href{https://lichess.org/@/maia1}{\texttt{maia1}} (Trained ELO 1100), \href{https://lichess.org/@/maia5}{\texttt{maia5}} (Trained ELO 1500) and \href{https://lichess.org/@/maia9}{\texttt{maia9}} (Trained ELO 1900).

\subsection{Move Prediction}\label{maia_sup}

\xhdr{Training}
To generate our training and validation sets, we considered all games played between January 2016 and November 2019 (we reserved December 2019 for the test sets only) and binned them into rating ranges of width 100.
To remove moves made under severe time pressure, we ignored games played at Bullet or HyperBullet time controls and removed all moves made in a game after either player had less than 30 seconds remaining.
We additionally wanted to ensure our training set covers the entire time range of our dataset.
To do so, we put games from each year (2017, 2018, 2019) into blocks of 200,000 games each, reserved the last three blocks from 2019 for use as validation sets, and randomly selected 20 blocks of games from each year.
This generated training sets of 12 million games and test sets of 200,000 games per rating bin.

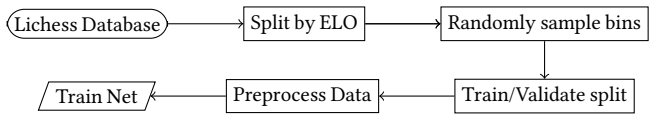
\begin{figure}[ht]
    \centering
    \small
    \tikzstyle{base} = [draw, align=center,minimum width=1cm,minimum height=.3cm]
    \tikzstyle{data} = [base, shape=rectangle,rounded corners=.8em]
    \tikzstyle{proc} = [base, shape=rectangle]
    \tikzstyle{net} = [base, trapezium, trapezium left angle=70, trapezium right angle=110]
    \tikzstyle{final} = [base, diamond]
    \begin{tikzpicture}
     \node [data]  (lichess) {Lichess Database};
     \node [proc, right =1cm of lichess] (split) {Split by ELO};
     \node [proc, right =1cm of split] (sample) {Randomly sample bins};
     \node [proc, below =.5cm of sample] (test) {Train/Validate split};
     \node [proc, left =1cm of test] (convert) {Preprocess Data};
     \node [net, left =1cm of convert] (net) {Train Net};
     \path[draw,->] (lichess) edge (split)
                    (split) edge (sample)
                    (split) edge (sample)
                    (sample) edge (test)
                    (test) edge (convert)
                    (convert) edge (net)
     ;
    \end{tikzpicture}
    \caption{Data Flow}
    \label{data_flow}
\end{figure}

Figure \ref{data_flow} shows how the data are processed.

\xhdr{Additional model information}
Our model architecture is comprised of 6 blocks of 2 CNNs each with 64 channels.
During training, moves are read sequentially from a game, but are only sampled with a probability of $\frac{1}{32}$.
During back propagation, both the policy head and value head losses are equally weighted, with the policy head using a cross entropy loss while the value head uses MSE.
We optimized all of our hyperparameters using our validation sets.

During training the sampled moves are put into a shufflebuffer with 250,000 slots before being used in a batch. Each model was trained in 400,000 steps using a batch size of 1024 and a learning rate of 0.01, preceded by a burn-in at .1 and succeeded by to drops by a factor 10, starting at step 200000. The final output of both heads goes from the tower of residual CNN blocks and into separate CNN layers followed by two fully connected layers for the policy head leading to a single number output, or 2 more CNN layers terminating in an output of $8\times8\times73$ which encodes the move. During testing only legal moves are considered from the output.

We chose to use 6 blocks and 64 filters, which was partially due to computational costs. Going to a larger network, such as 24 blocks and 320 filters, yielded a small performance boost at a significant cost in compute.

\begin{figure}[ht]
	\centering
	\includegraphics[width=.5\textwidth]{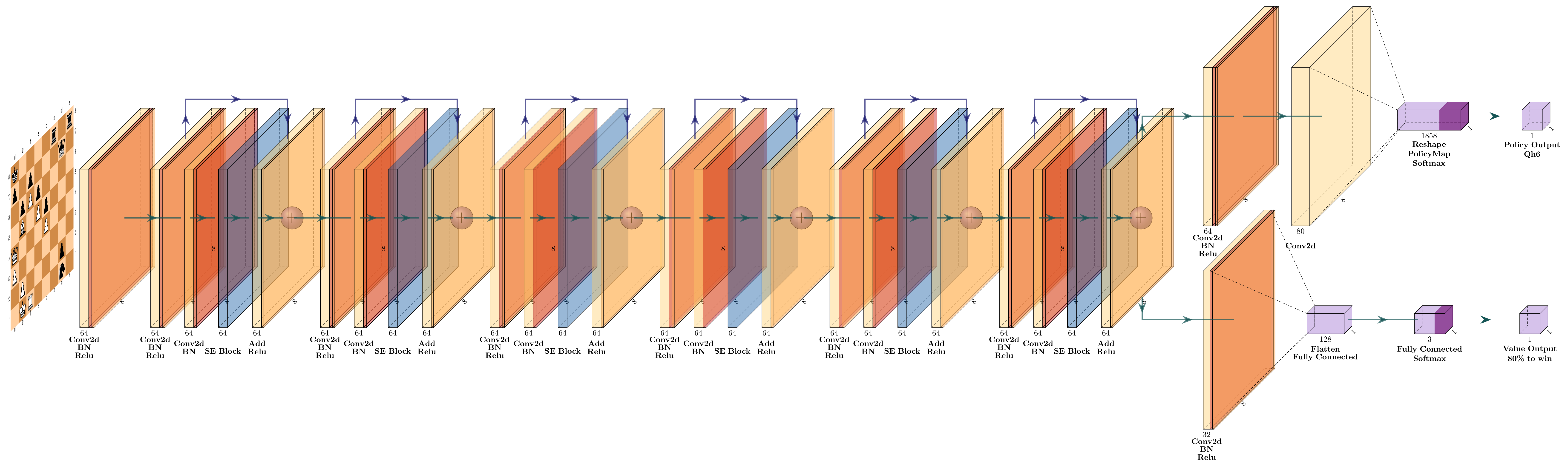}
	\caption{Visualization of the move prediction network}
	\label{network_diagram_maia_kdd}
\end{figure}

Figure \ref{network_diagram_maia_kdd} there is an an overview of the model in table \ref{maia_conf} has and the complete configuration files can be found on our code repository\footnotemark[\value{footnote}].

\begin{table}[ht]
    \centering
    \begin{tabular}{lr}
        \toprule
         Task & Move Prediction \\
         Model Type & Maia \\
         \midrule
         Blocks & 6\\
         Channels & 64\\
         Batch Size & 1024 \\
         Total Steps & 400000\\
         Initial Learning Rate & 0.1 \\
         Learning Rate Scaling Factor & .1 \\
         Learning Rate Scaling Steps & 80000, 200000, 360000 \\
         Optimizer & ADAM \\
         Framework & Tensorflow 2.0\\
         \bottomrule
    \end{tabular}
    \caption{Move prediction model configuration}
    \label{maia_conf}
\end{table}

\subsection{Centipawn to win probability}\label{cp_wr_supp}

\begin{figure}[ht]
	\centering
	\includegraphics[width=.5\textwidth]{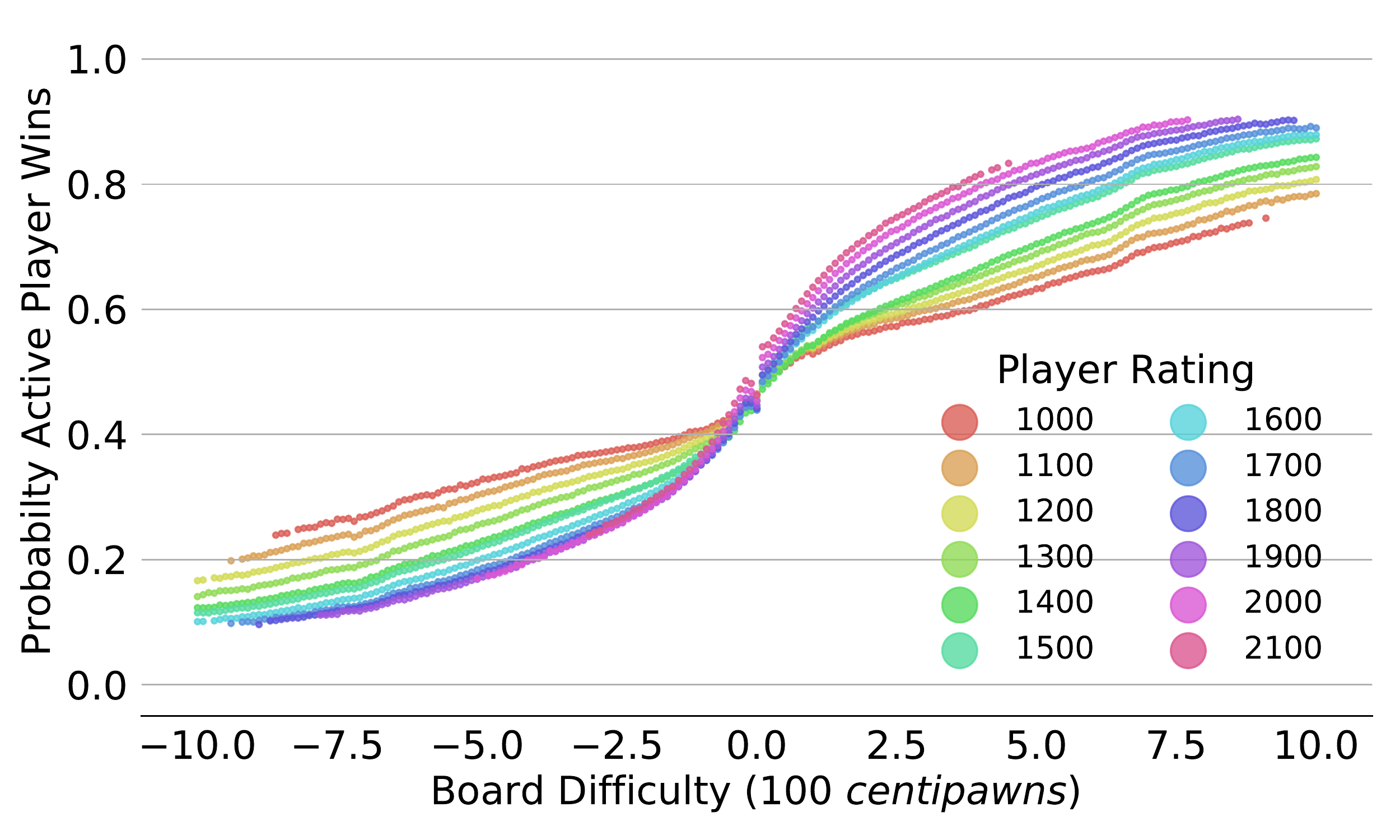}
	\caption{Conversion between board difficulty and empirical win probability, each point has 100,000 or more samples.}
	\label{CP_v_winrate_ELO}
\end{figure}

The conversion of centipawn score to win probability was done empirically by rounding to the nearest 10 centipawns (.1 pawns) and using the ratio wining players to the total number of observations at that value. Figure \ref{CP_v_winrate_ELO}, shows the distribution for different player ratings, but for our work we only looked at skill. The discontinuity near 0 is because a value of exactly 0 indicates following a loop is the optimal path for both players. Note that the starting board is rating as 20 centipawns in favour of white.

\subsection{Individual Blunder models}\label{blunder_sup}

\xhdr{Baseline Models}
The baseline models we tested are: decision tree, logit, linear regression, random forest and na\"{\i}ve bayes. We also tested SVM and perceptron models but they failed to finish in under 2 days. All models were from the Python library Scikit Learn.

\xhdr{Neural Models}
Boards were represented as a $8\times8\times17$ dimensional array with the 12 channels encoding pieces, 4 channels encoding castling, and one encoding whether the active player is white.

The residual CNN network has 6 residual blocks with two set of 2D CNNs with 64 channels and a $3\times3$ kernel. It was trained in the same way as the fully connected network, using an initial learning rate of 0.0002. Due to the small size of the input, the CNNs apply 0 padding to the inputs on all sides and use a stride length of 1 to make the output size equal to the input size.

The the configuration of both models used for individual blunder detection are shown in tables \ref{blunder_fc_conf} and \ref{blunder_resnet_conf}, for the fully connected and R-CNN models respectively.

Using the move prediction models for blunder prediction was as attempted but the results were far exceeded by the specialized models.

\begin{table}[ht]
    \centering
    \begin{tabular}{lr}
        \toprule
         Task & Move Prediction \\
         \multirow{2}{*}{Model Type} & Blunder Prediction \\
         & Fully Connected \\
         \midrule
         Layer sizes & 1028, 512, 256\\
         Activation Function & Hyperbolic Tangent\\
         Batch Size & 2000 \\
         Total Steps & 1400000\\
         Initial Learning Rate & 0.002 \\
         Learning Rate Scaling Factor & 0.1 \\
         Learning Rate Scaling Steps & 20000, 1000000, 1300000 \\
         Optimizer & ADAM \\
         Framework & Pytorch 1.3\\
         \bottomrule
    \end{tabular}
    \caption{Blunder prediction Fully connected model configuration}
    \label{blunder_fc_conf}
\end{table}

\begin{table}[H]
    \centering
    \begin{tabular}{lr}
        \toprule
         Task & Move Prediction \\
         \multirow{2}{*}{Model Type} & Blunder Prediction \\
         & Residual CNN \\
         \midrule
         Blocks & 6\\
         Channels & 64\\
         Batch Size & 2000 \\
         Total Steps & 1400000\\
         Initial Learning Rate & 0.0002 \\
         Learning Rate Scaling Factor & 0.1 \\
         Learning Rate Scaling Steps & 20000, 1000000, 1300000 \\
         Optimizer & ADAM \\
         Framework & Pytorch 1.3\\
         \bottomrule
    \end{tabular}
    \caption{Blunder prediction R-CNN model configuration}
    \label{blunder_resnet_conf}
\end{table}

\subsection{Grouped Blunder models}\label{grouped_sup}

We started with the same dataset as for individual behavior, then for every sample converted the board into a normalized form that only had piece locations and made the active player white. Then for each month we counted the number of games that encountered it and how many blunder were made per month for each normalized board. These were joined into the complete dataset if they had more than 1 samples in a month. Finally all boards with less than 10 samples total were discarded. The testing set was formed from a 10\% sample. The final training dataset has 1,066,055 boards, while the testing set has 118,582. One side effect of sampling only popular positions is that the middle game vanishes from our data.

The models used for the grouped blunder prediction task are the same as for the individual, but the fully connect model has an initial learning rate of 0.0001, while the R-CNN has an initial learning rate of 0.00001. The deep R-CNN also has 8 blocks and 256 filters. They also employ an early stopping criteria of 64 testing steps (done every 200 training steps) with no improvement in accuracy. Our final models were stopped before they fully memorized the data, as their accuracy on the holdout set started decreasing past 50 iterations. The final models stop at stop 378k, 248k and 248k respectively. The validation accuracy vs step is very consistent between the models, with the deep just being slightly higher.

The models used for this are the same architecturally to the individual task, but  normalization of the input boards causes the last 5 channels of the input board to be the same regardless of input. The balancing of positive and negative samples was also used.

\end{document}